\documentclass[10pt,twocolumn,letterpaper]{article}
\usepackage{iccv}
\usepackage{times}
\usepackage{epsfig}
\usepackage{graphicx}
\usepackage{amsmath}
\usepackage{amssymb}
\usepackage[breaklinks=true,bookmarks=false]{hyperref}
\usepackage{float}
\iccvfinalcopy 
\def\iccvPaperID{1690}

\ificcvfinal\fi
\usepackage{multirow}
\usepackage{colortbl}  
\usepackage{booktabs}
\usepackage{listings}
\usepackage{xparse}
\usepackage{marvosym}
\usepackage{lipsum}
\usepackage[table,xcdraw]{xcolor}

\definecolor{codegreen}{rgb}{0,0.6,0}
\definecolor{codegray}{rgb}{0.5,0.5,0.5}
\definecolor{codepurple}{rgb}{0.58,0,0.82}
\definecolor{backcolour}{rgb}{0.95,0.95,0.92}
\lstdefinestyle{mystyle}{
  backgroundcolor=\color{backcolour}, commentstyle=\color{codegreen},
  keywordstyle=\color{magenta},
  numberstyle=\tiny\color{codegray},
  stringstyle=\color{codepurple},
  basicstyle=\ttfamily\footnotesize,
  breakatwhitespace=false,         
  breaklines=true,                 
  captionpos=b,                    
  keepspaces=true,                 
  numbers=left,                    
  numbersep=5pt,                  
  showspaces=false,                
  showstringspaces=false,
  showtabs=false,                  
  tabsize=2
}

\def\consequence/{return}

\begin{document}
\title{Learning to Identify Critical States for Reinforcement Learning from Videos}
\author{
Haozhe Liu$^{1 \dagger}$, Mingchen Zhuge$^{1 \dagger}$, Bing Li$^{1 \textrm{\Letter}}$, Yuhui Wang$^1$, Francesco Faccio$^{1,2}$\\
Bernard Ghanem$^1$, Jürgen Schmidhuber$^{1,2,3}$ \\
$^1$AI Initiative, King Abdullah University of Science and Technology \\
$^2$The Swiss AI Lab IDSIA/USI/SUPSI, $^3$NNAISENSE \\
\texttt{\small \{haozhe.liu, mingchen.zhuge, bing.li, yuhui.wang,} \\
\texttt{\small francesco.faccio, bernard.ghanem, juergen.schmidhuber\}@kaust.edu.sa}
}

\maketitle
\ificcvfinal\thispagestyle{empty}\fi

\begin{abstract}

Recent work on deep reinforcement learning (DRL) has pointed out that algorithmic information about good policies can be extracted from offline data which lack explicit information about executed actions~\cite{schmidhuber2015learning,schmidhuber2018one,lecun2022path}. 
For example, videos of humans or robots may convey a lot of implicit information about rewarding action sequences, but a DRL machine that wants to profit from watching such videos must first learn by itself to identify and recognize relevant states/actions/rewards. Without relying on ground-truth annotations, our new method called Deep State Identifier learns to predict returns from episodes encoded as videos. Then it uses a kind of mask-based sensitivity analysis to extract/identify important critical states. Extensive experiments showcase our method's potential for understanding and improving agent behavior.
The source code and the generated datasets are available at \href{https://github.com/AI-Initiative-KAUST/VideoRLCS}{Github}.

\end{abstract}
\section{Introduction}

\let\thefootnote\relax\footnotetext{$\dagger$ Equal Contribution.}
\let\thefootnote\relax\footnotetext{$\textrm{\Letter}$ Corresponding Author.}
\let\thefootnote\relax\footnotetext{Accepted to ICCV23.}
In deep reinforcement learning (DRL), the cumulative reward---also known as the return---of an episode is obtained through a long sequence of dynamic interactions between an agent (i.e., a decision-maker) and its environment.  In such a setting, the rewards may be sparse and delayed, and it is often unclear which decision points were critical to achieve a specific return.

Several existing methods use the notion of localizing critical states, such as EDGE~\cite{guo2021edge} and RUDDER~\cite{arjona2019rudder}. These methods typically require explicit action information or policy parameters to localize critical states.
This limits their potential applicability in settings like video-based offline RL, where an agent's actions are often hard to measure, annotate, or estimate~\cite{zhu2023guiding,liu2018imitation}. To avoid this pitfall, in this work, we explicitly study the relationship between sequential visual observations and episodic returns without accessing explicit action information.

\begin{figure}[t!]
    \centering
    \includegraphics[width=0.475\textwidth]{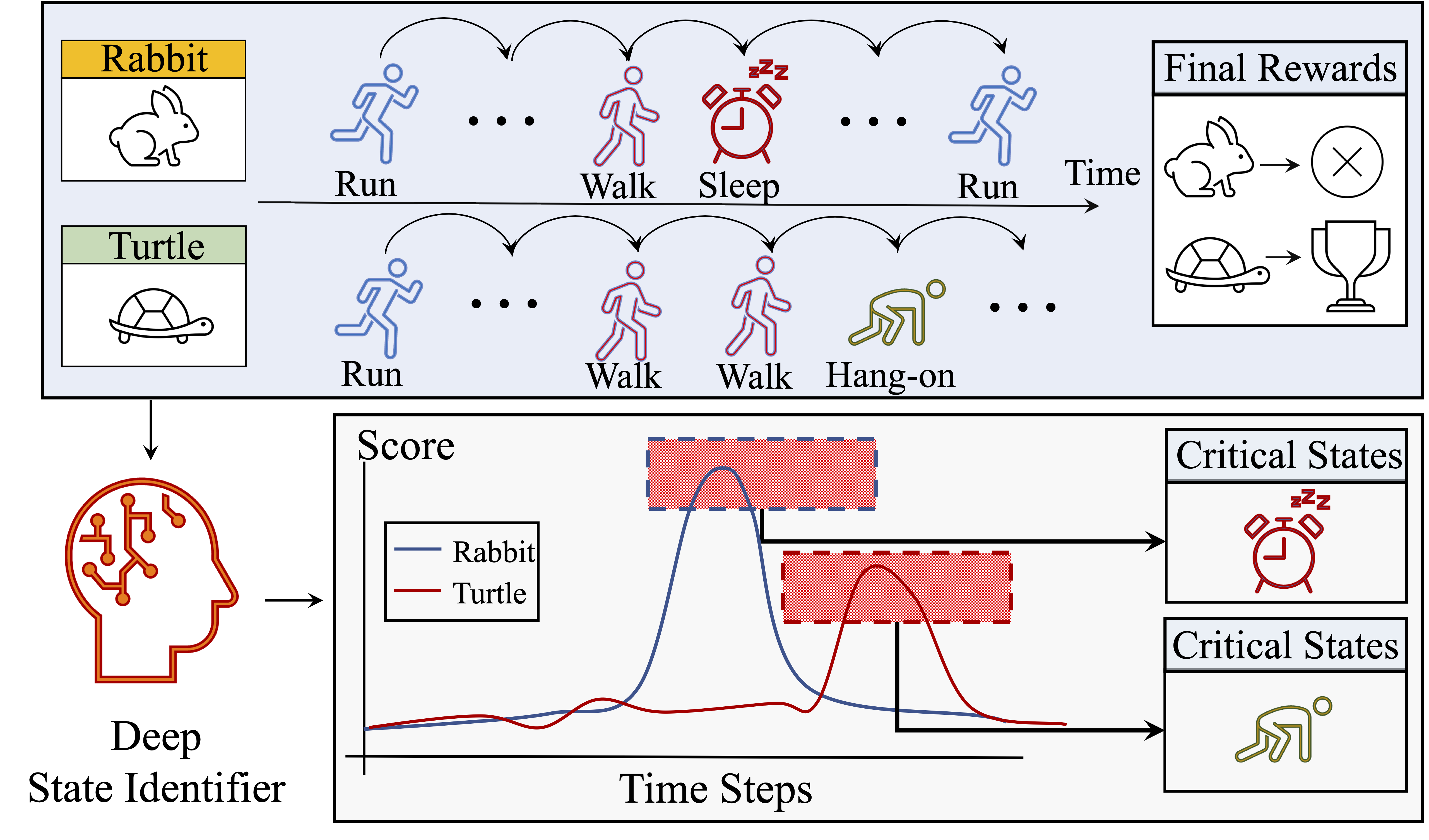}
 \caption{\textbf{Motivation of the proposed method.} In the illustrated race between a turtle and a rabbit, the \textit{sleep} state is critical in determining the winner of the race. Our method is proposed to identify such critical states. 
 }
    \label{fig:motivation}
\end{figure}
Inspired by the existing evidence that frequently only a few decision points are important in determining the return of an episode~\cite{arjona2019rudder,faccio2022general}, 
and as shown in Fig.~\ref{fig:motivation},
we focus on identifying the state underlying these critical decision points. However, the problem of directly inferring critical visual input based on the return is nontrivial~\cite{faccio2022general}, and compounded by our lack of explicit access to actions or policies during inference. To overcome these problems---inspired by the success of data-driven approaches~\cite{xie2022clims,nguyen2018weakly,hou2018self}---our method learns to infer critical states from historical visual trajectories of agents.

We propose a novel framework, namely the \emph{Deep State Identifier}, to identify critical states in video-based environments.
A principal challenge of working in such settings lies in acquiring ground-truth annotations of critical states; it is laborious to manually label in videos critical states corresponding to complex spatio-temporal patterns.
The Deep State Identifier is designed to directly overcome this challenge by identifying the critical states based solely on visual inputs and rewards. Our proposed architecture comprises a return predictor and a critical state detector. 
The former predicts the return of an agent given a visual trajectory, while the latter learns a soft mask over the visual trajectory where the non-masked frames are sufficient for accurately predicting the return. Our training technique explicitly minimizes the number of critical states to avoid redundant information through a novel loss function. If the predictor can achieve the same performance using a small set of frames, we consider those frames critical. Using a soft mask, we obtain a rank that indicates the importance of states in a trajectory, allowing for the selection of critical states with high scores. During inference, critical states can be directly detected without relying on the existence of a return predictor. Our contributions can be summarized as follows: 
\begin{itemize}
   \item We propose a novel framework that effectively identifies critical states for reinforcement learning from videos, despite the lack of explicit action information.
    \item  We propose new loss functions that effectively enforce compact sets of identified critical states.
    \item We demonstrate the utility of the learned critical states for policy improvement and comparing policies.
\end{itemize}

\section{Related Work}

In the past decade, researchers have explored the potential of combining computer vision (CV) and RL to develop more intelligent agents. A pioneering study by Koutnik et al.~\cite{koutnik2013evolving} used recurrent neural networks to tackle vision-based RL problems through an evolutionary strategy~\cite{koutnik2010evolving}. Since then, this topic has gained popularity. Mnih et al.~\cite{mnih2013playing, mnih2015human} trained a deep neural network using raw pixel data from Atari games to learn the Q-function for RL agents.
Recently, Visual MPC~\cite{finn2017deep} proposed a method using deep convolutional neural networks to predict the future states of a robot's environment based on its current visual input.
RIG~\cite{nair2018visual} trains agents to achieve imagined goals in a visual environment using a combination of RL and an auxiliary visual network. 
Ha and Schmidhuber~\cite{ha2018world} propose a version of the world model, which employs a Variational Autoencoder (VAE)~\cite{kingma2013auto} to construct representations of the visual environment and help train a model using imagined future states. Robotprediction~\cite{finn2016unsupervised} designs a method for unsupervised learning of physical interactions through video prediction, achieved by an adversarial model that assists RL agents in learning to interact with the environment. 
More recently, researchers have explored novel CV advances, such as self-attention and self-supervised learning, applied to RL algorithms~\cite{janner2021sequence,chen2021decision, yu2022mask,geng2022multimodal,eysenbach2022contrastive}, leading to satisfactory improvements. 
While visual input is integral to RL agents and can benefit RL in numerous ways, our paper proposes a method to assist agents in identifying the most crucial visual information for decision-making rather than solely focusing on improving visual representation.

Our method offers a novel perspective on explainable RL by identifying a small set of crucial states.
Explaining the decision-making process in RL is more challenging than in CV, due to its reliance on sequential interactions and temporal dependencies. 
Various methods have been employed to address this challenge.
Recent attention-based approaches~\cite{janner2021sequence,chen2021decision, mott2019towards} focus on modeling large-scale episodes offline~\cite{janner2021sequence,chen2021decision} to localize crucial decision-making points~\cite{mott2019towards}. 
However, the attention structure typically operates on feature space, where the spatial correspondence is not aligned with the input space~\cite{bau2020understanding, guo2021edge}. Therefore, it is challenging to directly threshold attention values to identify critical temporal points. Post-training explanation is an efficient method that directly derives the explanation from an agent's policy or value network~\cite{lu2021dance,guo2018lemna,guo2018explaining,fong2017interpretable}, thereby reducing memory and computation costs. 
Other popular explainable DRL methods include self-interpretable methods, such as Relational-Control Agent~\cite{zambaldi2018deep} and Alex~\cite{mott2019towards}, and model approximation methods, such as VIPER~\cite{bastani2018verifiable} and PIRL~\cite{verma2018programmatically}. These methods are widely used in the field of DRL~\cite{lu2021dance, guo2018lemna, guo2018explaining, fong2017interpretable, zambaldi2018deep, mott2019towards, bastani2018verifiable, verma2018programmatically}. 
For example, Alex~\cite{mott2019towards} proposes using the output of the attention mechanism to enable direct observation of the information used by the agent to choose its action, making this model easier to interpret than traditional models. 
Tang et al.~\cite{tang2020neuroevolution} use a small fraction of the available visual input and demonstrate that their policies are directly interpretable in pixel space. 
The PIRL method~\cite{verma2018programmatically} produces interpretable and verifiable policies using a high-level, domain-specific language.
Recent work uses policy fingerprinting~\cite{harb2020policy} to build a single value function to evaluate multiple DRL policies~\cite{faccio2022general,faccio2020parameter,faccio2022goal}.
The authors use only the policy parameters and the return to identify critical abstract states for predicting the return. However, policy parameters are often unavailable in practical applications, and storing them for multiple policies can require significant memory resources. We circumvent this issue by using visual states observed from the environment rather than relying on policy parameters. 

Apart from the methods mentioned above, reward decomposition is also popular. 
Such methods~\cite{shu2017hierarchical,juozapaitis2019explainable} re-engineer the agent's reward function to make the rewards earned at each time step more meaningful and understandable.
Compared to these methods, our approach evaluates the specific states. It provides a context-based framework for long-horizon trajectories in a challenging, yet practical domain, specifically learning without actions. Our method is also related to the concept of Hierarchical RL~\cite{hqlearning, sutton1999between}, which aims to identify high-level subgoals~\cite{schmidhuber1993planning, Schmidhuber:91chunker} that a low-level policy should achieve. 
Using a few crucial states to explain an RL agent is closely connected to the concept of history compression~\cite{Schmidhuber:91singaporechunker,jurgenlearningcomplex}, where a neural network is trained to learn compact representations that are useful for modeling longer data sequences.

\section{Method}
\subsection{Problem Formulation}
\NewDocumentCommand{\timestep}{ O{}} { ^{(#1)} }

In Reinforcement Learning (RL)~\cite{Sutton:2018:RLI:3312046}, an agent interacts sequentially with an environment. At each time step $t$, the agent observes a state $s^{(t)}$---in our case, the frame of a video, chooses an action $a^{(t)}$, obtains a scalar immediate reward $r^{(t)} = R(s^{(t)}, a^{(t)})$, where $R$ is the reward function, and transitions to a new state $s^{(t+1)}$ with probability $P(s^{(t+1)} |s^{(t)}, a^{(t)})$. 

The behavior of an agent is expressed by its policy $\pi(a|s)$, which defines a probability distribution over actions given a state.
The agent starts from an initial state and interacts with the environment until it reaches a specific state (a goal state or a failing state) or hits a time horizon $T$. 
Each of these interactions generates an \textit{episode} and a \emph{return}, i.e., the discounted cumulative reward $\mathbf{y} = \sum_{t=0}^{T}{\gamma^t r^{(t)} }$, where $\gamma \in [0,1)$ is a discount factor. 
Due to the general form of the return and the complex agent-environment interaction, it is generally difficult to identify which decision points---or states---are essential to achieve a specific return in an episode. In other words, it is difficult to explain the behavior of a policy. 

Inspired by the success of data-driven approaches~\cite{xie2022clims,nguyen2018weakly,hou2018self,zhuge2022salient}, we design a learning-based method to identify a few crucial states in an episode that are critical to achieving the return $\mathbf{y}$. Unlike previous approaches~\cite{arjona2019rudder,guo2021edge}, we focus on identifying critical states in a video without needing an explicit representation of the policy or actions executed.
More formally, let $\{\mathbf{s}_i, \mathbf{y}_i\}_i$ be the collected \emph{episode-return training data}, where $\mathbf{s}_i=\{s_i\timestep[t]\}_{t}$ is the $i$-th state trajectory, $s_i\timestep[t]$ is a state at the time step $t$, and $\mathbf{y}_i$ is the return achieved in the state trajectory $\mathbf{s}_i$. 

\begin{figure}[tbp]
    \centering
    \includegraphics[width=0.475\textwidth]{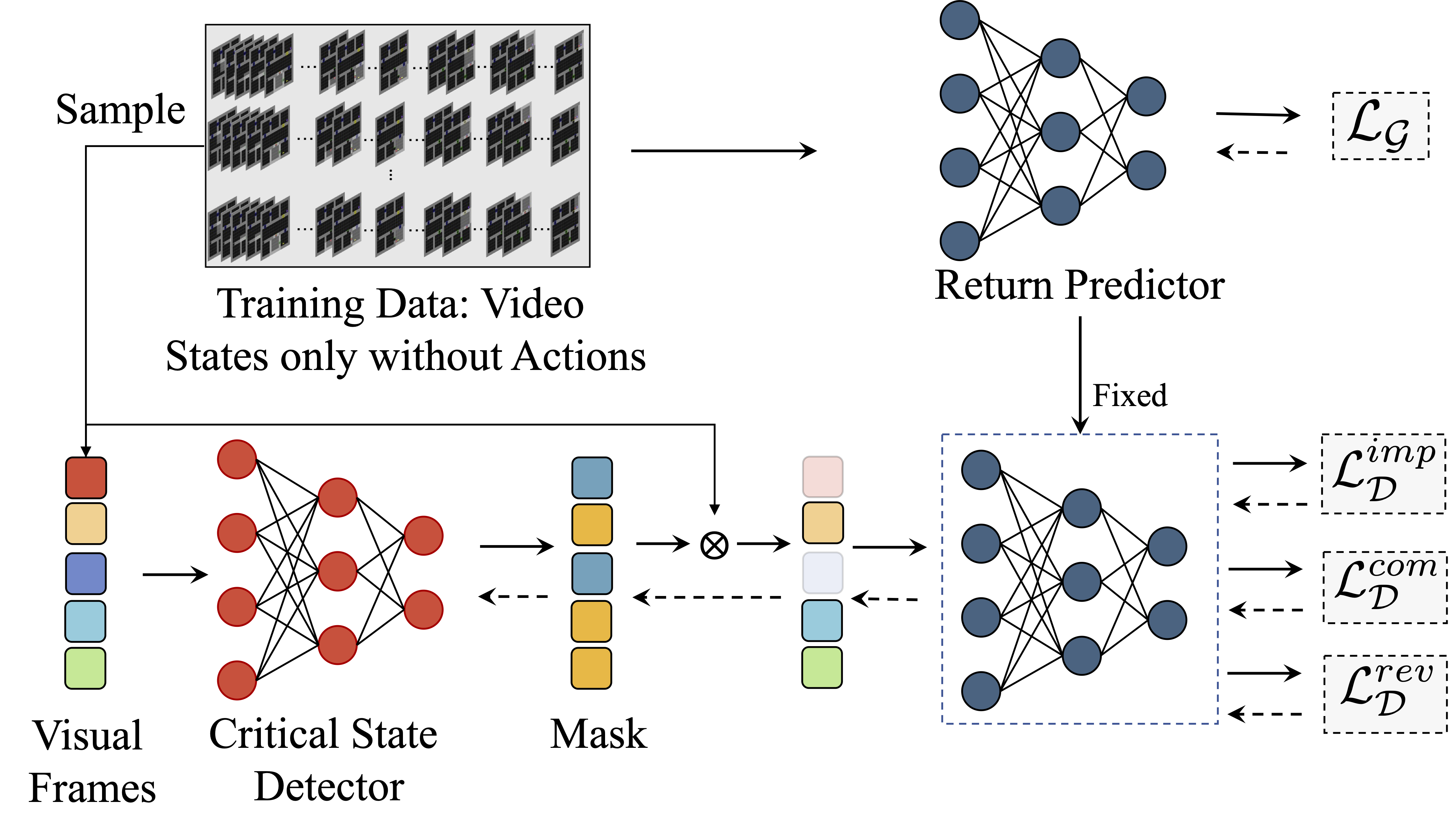}
   \caption{\textbf{Illustration of the proposed framework.} 
   During training, our return predictor learns to predict the return of an episode from a state trajectory. Our critical state detector learns to exploit the return predictor to identify a compact set of states critical for return prediction. During testing, the critical state detector takes a state trajectory as input and automatically detects its critical states without using the return predictor.}
   \label{fig:general}
\end{figure}

To identify critical states, we suggest a novel framework, called the Deep State Identifier, consisting of the following two steps.
\textbf{First}, we propose a return predictor that estimates the return $\mathbf{y}_i $ given a state trajectory $\mathbf{s}_i$. \textbf{Second}, we use the return predictor to train a critical state detector to identify critical states. The detector receives the states as input and outputs a mask over the states. It is used to measure how important each state is to the return. Fig.~\ref{fig:general} illustrates the architecture of our method.  

\subsection{Return Predictor}\label{sec:te}
  
Our return predictor $\mathcal{G}(\cdot)$ aims to predict the return of a sequence of states. We build it using a neural network and train it in a supervised manner.
There are two types of learning objectives depending on whether the return is discrete or continuous. 
For discrete return values (e.g., $1$ indicates success, while $0$ denotes failure), we train $\mathcal{G}(\cdot)$ using cross-entropy loss:
\begin{align} 
\mathcal{L}^c_\mathcal{G} =\sum_{i} \mathcal{L}^c_\mathcal{G} (\mathbf{s}_i,\mathbf{y}_i) = -\sum_{i}\mathbf{y}_i log\mathcal{G}(\mathbf{s}_i), 
\end{align}
where $\mathbf{y}_i $ is the category-level annotation of ${s}_i$.
If the return is continuous,
we employ a regression loss $\mathcal{L}_\mathcal{G}^r$ to train   $\mathcal{G}(\cdot)$, 
\begin{align}
\mathcal{L}^r_\mathcal{G}=\sum_i \mathcal{L}_\mathcal{G}^r(\mathbf{s}_i,\mathbf{y}_i) = \sum_{i}||\mathcal{G}(\mathbf{s}_i) - \mathbf{y}_i ||_2,
\end{align}
where $\mathbf{y}_i \in \mathbb{R} $ is the scalar return of state trajectory $\mathbf{s}_i$.

\subsection{Critical State Detector}
\label{sec:sd}

In a general environment, manually labeling critical states is expensive and impractical. The unavailability of ground-truth critical states prevents our method from being fully-supervised.
We hereby propose a novel way of leveraging the return predictor for training a critical state detector. Note that the critical states are elements of the state trajectory and can be discontinuous along the temporal dimension. 
We cast the task of identifying critical states as deriving a soft mask on a state trajectory. In particular, given a state trajectory $\mathbf{s}_i=\{s^{(t)}_i\}$, the critical state detector $\mathcal{D}$ outputs a mask on $\mathbf{s}_i$, \ie, $\mathbf{m}_i=\mathcal{D}(\mathbf{s}_i)$, where $\mathbf{m}_i=\{m^{(t)}_i\}$, $m^{(t)}_i\in [0~1]$ can be interpreted as confidence that $s^{(t)}_i$ is a critical state. Intuitively, a high value of $m^{(t)}_i$ indicates a higher probability that the corresponding state $s_i^{(t)}$ is critical.
To enforce $\mathcal{D}$ to  identify critical states, we design three loss functions, namely, importance preservation loss,  compactness loss, and reverse loss, for training  $\mathcal{D}$:

\begin{align}
    \mathcal{L}_\mathcal{D} = \lambda_s \mathcal{L}^{imp}_{\mathcal{D}} + \lambda_r \mathcal{L}_\mathcal{D}^{com} + \lambda_v \mathcal{L}_\mathcal{D}^{rev},
\end{align}
where $\lambda_s$, $\lambda_r$ and $\lambda_v$ are the weights for {importance preservation loss}, {compactness loss}, and {reverse loss} respectively.

\noindent \textbf{Importance preservation loss.} Given a state trajectory $\mathbf{s}_i$, the goal of the importance preservation loss is to ensure the states discovered by the critical state detector are important to predict the return $\mathbf{y}_i$.  
Hence, the loss enforces the masked state sequence discovered by $\mathcal{D}$ to contain a similar predictive information of the original state trajectory  $\mathbf{s}_i$. Given the training data $\{(\mathbf{s}_i, \mathbf{y}_i)\}$, the importance preservation loss is defined as follows: 

\begin{align}
    \mathcal{L}^{imp}_{\mathcal{D}} =\sum_{i} \mathcal{L}_\mathcal{G}(\mathcal{G}(\mathbf{s}_i \circ \mathcal{D}(\mathbf{s}_i)),\mathbf{y}_i),
\end{align}
where $\circ$ denotes the element-wise multiplication $(\mathbf{s}_i \circ \mathcal{D}(s_i))^{(t)} \triangleq m_i^{(t)} {s}_i^{(t)} $ , $\mathcal{G}(\mathbf{s}_i \circ \mathcal{D}(\mathbf{s}_i))$ predicts the return of the masked state sequence $\mathbf{s}_i \circ \mathcal{D}(\mathbf{s}_i)$,  $\mathcal{L}_\mathcal{G}$ stands for $\mathcal{L}^c_\mathcal{G}$ or  $\mathcal{L}^r_\mathcal{G}$, as defined in the previous subsection. Note that the masked state sequence can be discontinuous, and the information is dropped by skipping some redundant states. As a result, we cannot obtain a ground-truth return for a masked state sequence by running an agent in its environment.
Thanks to the generalization abilities of 
neural networks~\cite{zhou2022domain,wang2022generalizing,schmidhuber2022annotated,schmidhuber2015deep}, we expect that the return predictor trained on the original state trajectories can predict well the return for masked state trajectories when critical states are not masked. 

\noindent \textbf{Compactness loss.} Solely using the importance preservation loss $\mathcal{L}^{imp}_{\mathcal{G}}$ leads to a trivial solution where the mask identifies all states in $\mathbf{s}_i$ as critical. 
Critical states should instead be as compact as possible to avoid involving redundant and irrelevant states. 
To address this issue, we further introduce the compactness loss $\mathcal{L}_\mathcal{D}^{com}$. The compactness loss forces the discovered critical state to be as few as possible. 
Specifically, we employ the L1-norm to encourage  the mask, \ie, the output of $\mathcal{D}$, to be sparse given each $\mathbf{s}_i$ : 
\begin{align}
    \mathcal{L}_\mathcal{D}^{com} = \sum_{i}  || \mathcal{D}(\mathbf{s}_i)||_1.
\end{align}
It is difficult to balance the importance preservation loss and compactness loss. The detector may ignore some critical states for compactness.  
We propose a 
reverse loss for training $\mathcal{D}$ to mitigate this problem.

\noindent \textbf{Reverse loss.} The third loss is designed for undetected states. 
We remove the critical states by inverting the mask from the original state trajectory $\mathbf{s}_i \circ (1 - \mathcal{D}(\mathbf{s}_i))$ and process this masked sequence where the remaining states are useless for return prediction. 
This loss ensures that all the remaining states are not useful for estimating the return. 
We define the reverse loss as:

\begin{align}
    \mathcal{L}_\mathcal{D}^{rev} = -\sum_{i}\mathcal{L}_\mathcal{G} (\mathcal{G}(\mathbf{s}_i \circ (1-\mathcal{D}(\mathbf{s}_i))), \mathbf{y}_i).
\end{align}

\subsection{Iterative Training}

Here we introduce the training strategy of our framework.    
We train the return predictor on complete and continuous state trajectories. At the same time, we use it to predict the return of masked state sequences that are incomplete and discontinuous when training the critical state detector.  
 We iteratively train the predictor and the detector,  where the learning objective of the whole framework is given by: 
\begin{align}
    \min_\mathcal{G}\min_\mathcal{D}   \mathcal{L}_\mathcal{D} + \mathcal{L}_\mathcal{G}.
\end{align}
After training, our critical state detector automatically detects critical states without using the return predictor. Appendix A lists the pseudo-code of the proposed method. 

\section{Experiments}

\subsection{Benchmark and Protocol Navigation}
We begin this section by releasing a benchmark to test our method and facilitate the research on explainability. As shown in Table \ref{tab:collected_dataset}, we collect five datasets on three different RL environments, i.e., Grid World~\cite{minigrid,chevalier2018babyai}, Atari-Pong~\cite{1606.01540}, and Atari-Seaquest~\cite{1606.01540}. We select Grid World for qualitative analysis since it is very intuitive for human understanding.
We study a challenging environment with partial observation.
In the context of Grid World, we define a "state" as a combination of the current visual frame and historical information. Although this surrogate representation does not equate to the full, true state of the environment, it serves as an agent's internal understanding, developed from its sequence of past observations. To elaborate, when we say that our model identifies a "state" in this context, we imply that it recognizes a specific observation or frame, based on the agent's history of previous observations.
For fully observable environments like Atari, the term "state" assumes its traditional definition, providing complete information about the system at any given time.
We use Atari-Pong and Atari-Seaquest environments to compare our method with similar approaches based on critical state identification, using adversarial attacks, and evaluating policy improvement. Note that evaluating critical states using adversarial attacks was first proposed by work on Edge~\cite{guo2021edge}. However, Edge does not consider cross-policy attacks where the policies for training and testing the detector are different.   
More details can be found in the supplementary material.

\begin{table}[]
\caption{\textbf{The specification of the five collected datasets.} The datasets cover discrete and continuous returns for a comprehensive study of the proposed method. $\mathbf{y}$ here is the cumulative reward.}
\label{tab:collected_dataset}
\resizebox{.475\textwidth}{!}{
\begin{tabular}{lcccc}
\hline
\multicolumn{1}{c|}{}                 & \multicolumn{1}{c|}{Length} & Training    & \multicolumn{1}{c|}{Test}      & Total       \\ \hline
\multicolumn{5}{c}{\cellcolor[HTML]{EFEFEF}Grid World-S (Memory: 353 MB)}                                                                               \\ \hline
\multicolumn{1}{l|}{Reaching Goal}    & \multicolumn{1}{c|}{31.97}        & 1000        & \multicolumn{1}{c|}{200}       & 1200        \\ \hline
\multicolumn{1}{l|}{Fail}             & \multicolumn{1}{c|}{25.72}        & 1000        & \multicolumn{1}{c|}{200}       & 1200        \\ \hline
\multicolumn{5}{c}{\cellcolor[HTML]{EFEFEF}Grid World-M (Memory: 412 MB)}                                                                               \\ \hline
\multicolumn{1}{l|}{Policy-1}         & \multicolumn{1}{c|}{31.97}             & 1000        & \multicolumn{1}{c|}{200}       & 1200        \\ \hline
\multicolumn{1}{l|}{Policy-2}         & \multicolumn{1}{c|}{38.62}             & 995        & \multicolumn{1}{c|}{200}       & 1195        \\ \hline
\multicolumn{5}{c}{\cellcolor[HTML]{EFEFEF}Atari-Pong-{[}S/M{](Memory: 174 GB /352 GB)}}                                                                       \\ \hline
\multicolumn{1}{l|}{Agent Win}        & \multicolumn{1}{c|}{200}          & 13158/17412 & \multicolumn{1}{c|}{1213/1702} & 14371/19114 \\ \hline
\multicolumn{1}{l|}{Agent Lose}       & \multicolumn{1}{c|}{200}          & 8342/4088   & \multicolumn{1}{c|}{787/298}   & 9129/4386   \\ \hline
\multicolumn{1}{l|}{Total}            & \multicolumn{1}{c|}{-}            & 21500       & \multicolumn{1}{c|}{2000}      & 23500       \\ \hline
\multicolumn{5}{c}{\cellcolor[HTML]{EFEFEF}Atari-Seaquest-S (Memory:706 GB)}                                                                             \\ \hline
\multicolumn{1}{l|}{$\mathbb{E}[\mathbf{y}]$=2968.6}
& \multicolumn{1}{c|}{2652.5}       & 8000        & \multicolumn{1}{c|}{2000}      & 10000       \\ \hline
\end{tabular}
}
\end{table}

\begin{table}[]
\caption{\textbf{Summary of improvements due to our method},
where Gain refers to improvement over the baselines. Our method improves performance across various tasks.   The baselines in the 2nd-6th rows are our method using \textit{Imp. Loss} on Grid-World-S, EDGE \cite{guo2021edge} for Atari-Pong-S, an attack with 30 randomly selected frames on Atari-Pong-M, and DQN trained with 25M time steps on Atari-Seaquest-S, respectively.}
\vspace{5pt}
\label{tab:sum}
\resizebox{.49\textwidth}{!}{
\Large
\footnotesize
\begin{tabular}{c|c|l|>{\columncolor[HTML]{EFEFEF}}c }
\toprule
Datasets           & Navigation          & \multicolumn{1}{c|}{Task} & \textbf{Gain}        \\ \midrule
GridWorld-S        & Sec. \ref{sec:expl} & Critical State Identify   & \textbf{16.38\%}                    \\ 
GridWorld-S        & Sec. \ref{sec:expl} & Sequence Reasoning        & \textbf{Qualitative}                    \\ 
GridWorld-M        & Sec. \ref{sec:comp} & Policy Evaluation         & \textbf{First Study} \\ 
Atari-Pong-S       & Sec. \ref{sec:atta} & In-Policy Adv. Attack     & \textbf{18.63\%}     \\ 
Atari-Pong-M       & Sec. \ref{sec:atta} & Robust Analysis  & \textbf{50.35\%}     \\ 
Atari-Seaquest-S   & Sec. \ref{sec:impr} & Policy Improvement        & \textbf{17.65\%}      \\ \bottomrule
\end{tabular}
}
\end{table}
\begin{figure}[tbp]
    \centering
    \includegraphics[width=0.49\textwidth]{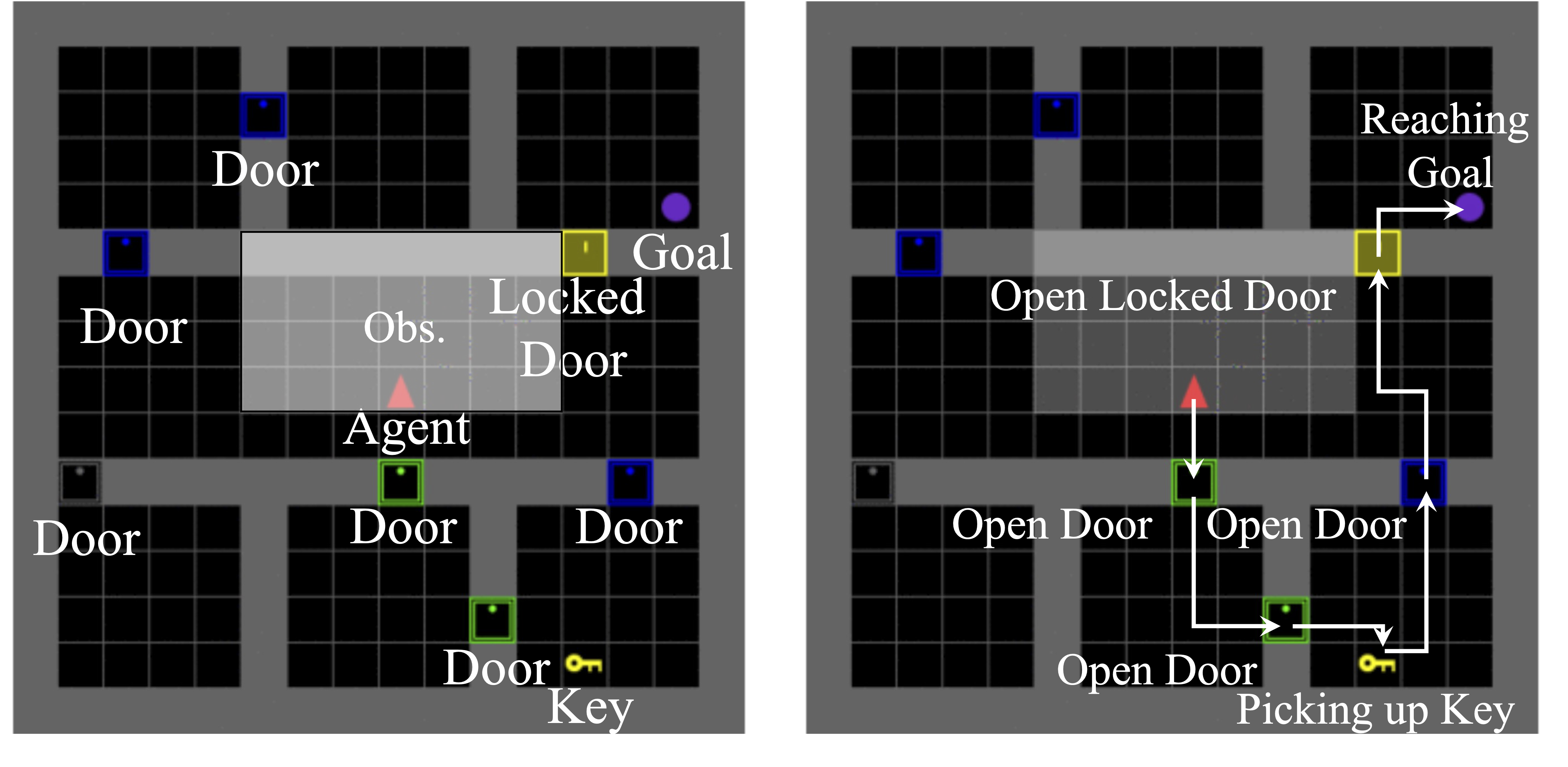}
   \caption{\textbf{Illustration of an instance of the GridWorld environment.} The environment consists of six rooms and one corridor. The agent starts from a random initial location in the corridor, and the final goal is to get the ball. Since the ball is locked in a room, the agent must pick up the key and open the yellow locked door. In a successful episode, the agent must open the unlocked doors (colored in green and blue), pick up the key, unlock the yellow door, and reach the purple ball. Note that the agent has only partial observation  (colored white)  of the environment at a time step.}
   \label{fig:dataset}
\end{figure}

\begin{figure*}[!h]
    \centering
    \includegraphics[width=0.98\textwidth]{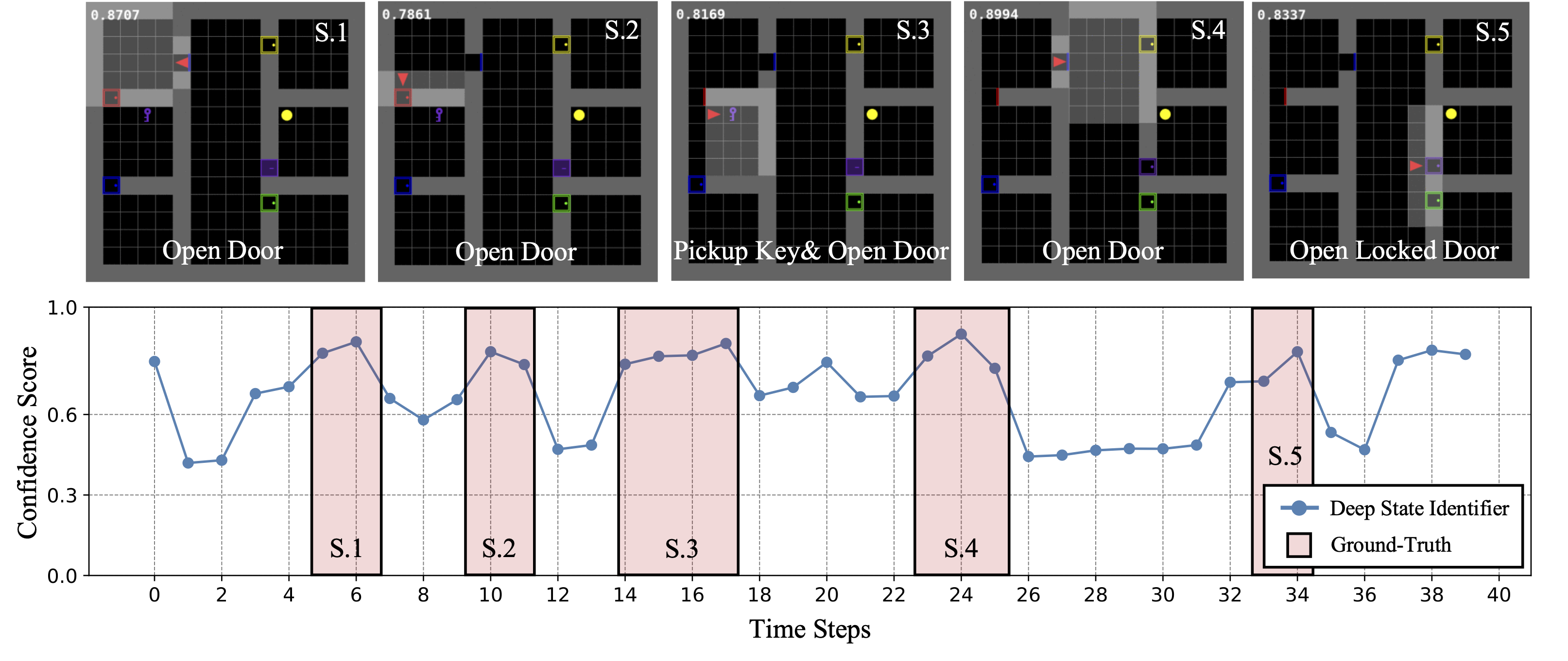}
   \caption{\textbf{The performance of our method in identifying critical states.} The top row shows human-annotated critical states (i.e., ground truth) in an episode. The bottom row shows for each time step in the environment how confident the detector is that the current state is critical. Our method assigns high scores to human-annotated critical states, demonstrating its identification abilities.}
   \label{fig:ours_visualization}
   \vspace{-1em}
\end{figure*}

\subsection{Critical State Discovery}
\label{sec:expl}
\noindent \textbf{Performance.} This section provides a qualitative analysis of the critical time point identified by our Deep State Identifier. We choose the `MiniGrid-KeyCorridorS6R3-v0' task~\cite{schmidhuber1996simple,minigrid} of the GridWorld environment, where the goal is to reach a target position in a locked room after picking up a key (see the yellow block in Fig.~\ref{fig:dataset}). This task is useful to visually evaluate our method since it is intuitive to identify what states are critical: top row in Fig.~\ref{fig:ours_visualization} shows that states immediately before actions such as `opening the door' (S.1, S.2, S.3 ), `picking up the key' and 'opening the locked door' are critical to successfully reaching the goal. 
Note that there is no ground truth on the critical state for a general, more complex environment.

We use a pre-defined DRL agent to collect trajectories. Since our method detects critical states by masking the trajectory, we evaluate how our critical state detector accurately assigns high scores to the states we intuitively labeled as critical. As shown in Fig.~\ref{fig:ours_visualization}, our method assigns high values to human-annotated critical states and low values to remaining states, showing its effectiveness in discovering critical states.   

\begin{table}[tbp]
\caption{\textbf{Ablation study for the critical state detector.} 
}
\vspace{5pt}
\label{tab:ablation_study}
\setlength\tabcolsep{16pt}
\resizebox{.475\textwidth}{!}{
\begin{tabular}{c|c|c|c}
\hline
\textit{Imp. Loss}    & \textit{Com. Loss}     & \textit{Rev. Loss}    & F-1 Score (\%)$\uparrow$ \\ \midrule
$\checkmark$ & $\times$     & $\times$         & 68.98          \\ 
$\checkmark$ & $\checkmark$ & $\times$         & unstable       \\ 
$\times$ & $\checkmark$ & $\checkmark$         & 74.42       \\ 
$\checkmark$ & $\times$     & $\checkmark$     & 76.09          \\ 
\rowcolor[HTML]{EFEFEF} 
$\checkmark$ & $\checkmark$ & $\checkmark$ & \textbf{80.28} \\ \bottomrule
\end{tabular}
}
\end{table}
\begin{figure}[!h]
    \centering
    \includegraphics[width=0.49\textwidth]{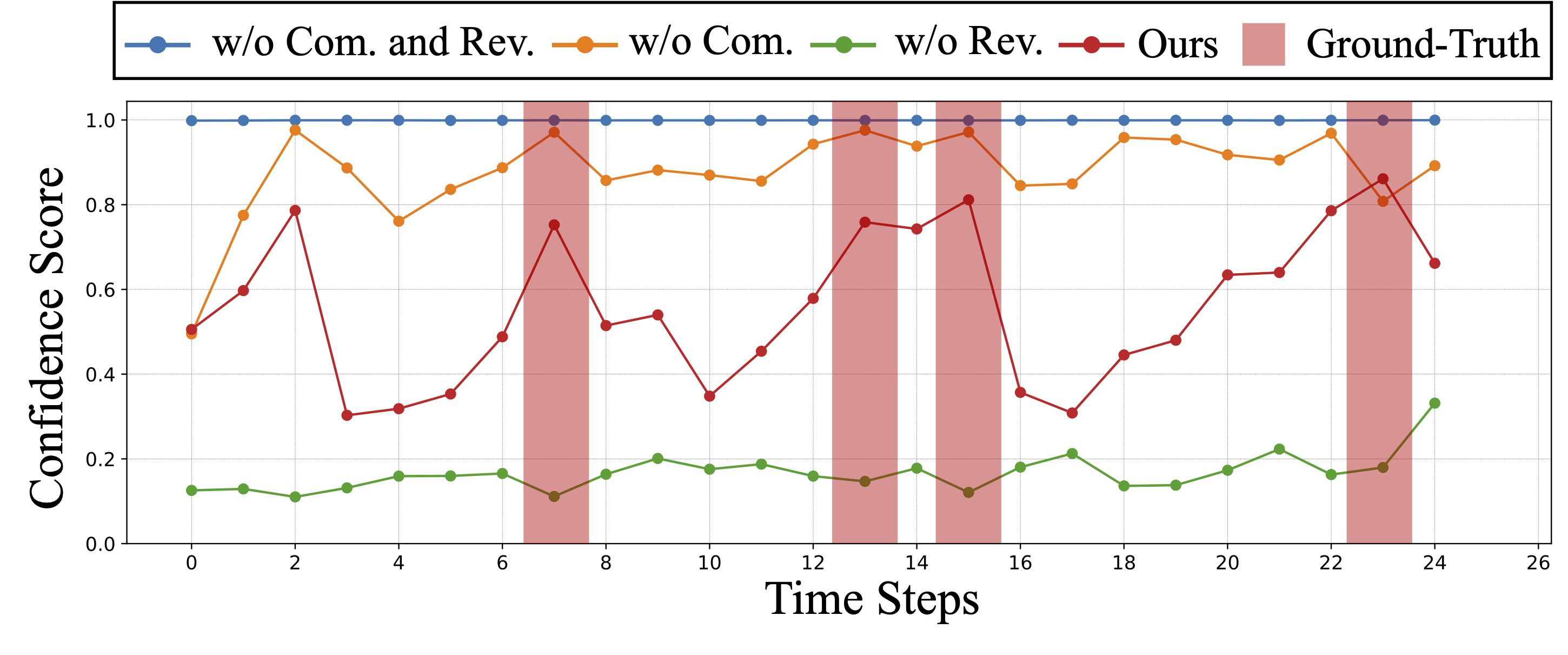}
    \vspace{-1em}
   \caption{\textbf{Ablation study of the detector's loss function.} For each time step and loss component, the line indicates how confident the detector is that the current input is critical. Red blocks mark the human annotation.}
   \vspace{-1em}
   \label{fig:ablation_study} 
\end{figure}

\noindent \textbf{Ablation study.} We analyze the contribution of each component of the critical state detector loss in Tab. \ref{tab:ablation_study}  and Fig.~\ref{fig:ablation_study}.
If we remove the compactness loss and the reverse loss, our method wrongly assigns high confidence to all states in an episode, \ie, all states are detected as critical ones. 
Similarly, if we remove the reverse loss, our method detects all states as non-critical. Finally, removing only the compactness loss, most states (including non-critical ones) are wrongly detected as critical.
This ablation shows that each loss component is crucial to critical state identification.  

\let\thefootnote\relax\footnotetext{$^1$We use a text description of states due to space constraints. We provide visual states in the supplemental material.\nobreak}

\noindent \textbf{More Analysis.} In RL, states within an episode can be highly correlated. We show how our method can discover state dependencies essential to identifying critical states. It is challenging to capture the dependencies among states in the Gridworld since the agent can only partially observe the environment through a small local view. 

Tab.  \ref{tab:sequence_reasoning} provides examples of states in the environment$^1$.
In Gridworld, the states that occur immediately before or after the action ``opening door" are frequently observed in a trajectory. In these states, the agent can be either with or without the key. However, obtaining the key is crucial for achieving the goal of GridWorld (see Fig.~\ref{fig:dataset}). Without the key, the agent cannot successfully finish the task. Therefore, the states immediately before or after the action ``opening door" without the key are not as critical as the states immediately before or after the action ``opening the door" with the key to predict the return.
Tab. \ref{tab:sequence_reasoning} shows how our method captures such dependencies between ``opening door" and ``picking up the key." Our method successfully assigns much higher confidence to the critical states immediately before or after the action ``opening door" with the key and lower confidence to the states immediately before or after the action ``opening door" without the key. 

\begin{table}[tbp]
\caption{\textbf{State detector's confidence score over different states.} Our method has different confidence scores for the states immediately before and after (i.b.a.) opening a door with or without the key, which indicates that it can capture temporal dependencies among states. Normal states refer to states where the agent has a distance greater than two from positions where it can take a relevant action (pick up the key or open a door). We report the mean and standard deviation of the confidence over four random seeds.}
\vspace{5pt}
\label{tab:sequence_reasoning}
\footnotesize
\centering
\begin{tabular}{p{5.9cm}|>{\centering\arraybackslash}p{1.5cm}}
\toprule
State Description                              & Confidence Score \\ \hline
Normal States (Full)                     &  53.66 $\pm$ 0.12                \\ \hline 
Normal States Before Picking up the Key  &  49.59 $\pm$ 0.13                \\ 
\rowcolor[HTML]{EFEFEF} 
State i.b.a. Opening Door (without the Key)      &  \textbf{67.13 $\pm$ 0.12}                \\  
\rowcolor[HTML]{C0C0C0}
State i.b.a. Trying Locked Door (without the Key)&   50.81 $\pm$  0.08                            \\ \hline
State i.b.a. Picking up the Key                       &  \textbf{78.35 $\pm$ 0.04}                \\ \hline 
Normal States After Picking Up the Key   &  56.58 $\pm$ 0.10                \\
\rowcolor[HTML]{EFEFEF} 
State i.b.a. Opening Door (with the Key)       &  \textbf{80.65 $\pm$ 0.06}                \\ 
\rowcolor[HTML]{C0C0C0}
State i.b.a. Opening Locked Door                         &  \textbf{87.55 $\pm$ 0.01}                \\ \bottomrule
\end{tabular}
\vspace{-2em}
\end{table}

\begin{figure*}[!h]
    \centering
    \includegraphics[width=0.98\textwidth]{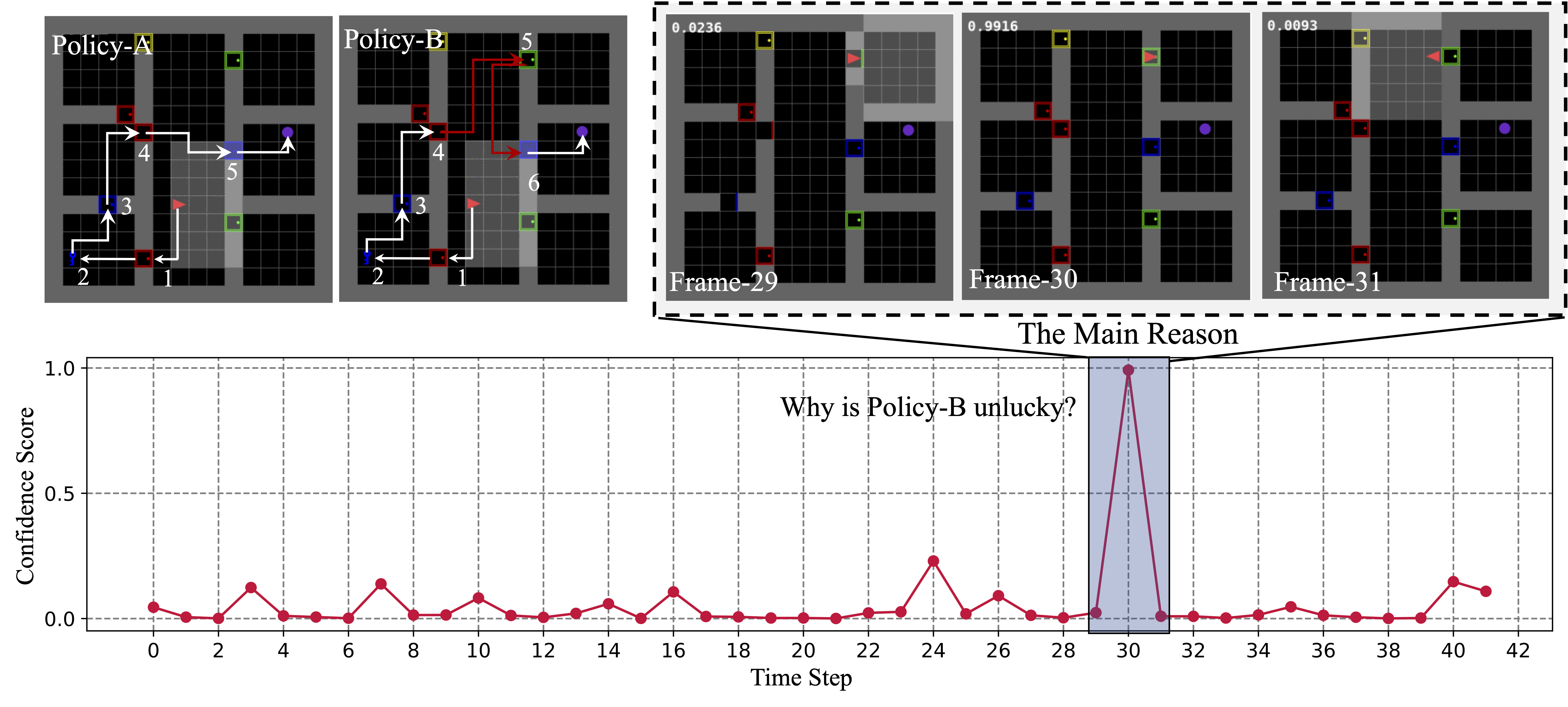}
   \caption{\textbf{Visualization of the Deep State Identifier for policy comparison.} We pre-collect policy-\textit{A} and policy-\textit{B}. While policy-\textit{A} is optimal, policy-\textit{B} first causes the agent to enter the incorrect room after picking up the key and then reach the goal. We train our method to discriminate between policy-\textit{A} and policy-\textit{B}, given sequences of trajectories generated by them. The critical state detector assigns high confidence to states where policy-\textit{B} is suboptimal.
   }
   \vspace{-1.4em}
   \label{fig:evaluation}
\end{figure*}

\subsection{Policy Comparison by Critical States}
\label{sec:comp}

In general, researchers use cumulative rewards to validate policy performance. However, these metrics cannot elucidate the diverse behavioral patterns exhibited by different policies. 
To better distinguish and explain the behavioral differences among various policies, a return predictor is trained to recognize the distinct trajectories of each policy. Our detector then is trained to identify critical states for highlighting the contrasts between policies rather than merely focusing on returns, thus facilitating a more comprehensive comparison of their behaviors.
Consequently, we can leverage the ability of the critical state detector to pinpoint the key states that discriminate between the two policies and visually represent the dissimilarities between them. As shown in Fig.~\ref{fig:evaluation}, both policy-\textit{A} and policy-\textit{B} can achieve the final goal, but in policy-\textit{B}, the agent always enters an invalid room after picking up the key, leading to more steps in the environment before achieving the goal. Both policies achieve a high return. However, our approach identifies the most discriminating states. Our method precisely assigns the highest confidence to the states inside the invalid room. The visualization shows that our method can explain the difference between the two policies. More details are provided in Appendix A. 

\subsection{Efficient Attack using Critical States}
\label{sec:atta}
In the previous sections, we showed that our method identifies the critical states with the highest impact on return prediction. However, for complex environments, it is difficult to evaluate the performance of this class of methods because the ground-truth critical states are not available. Following previous approaches~\cite{guo2021edge}, we use adversarial attacks to validate whether the identified states are critical. Intuitively, if a state is critical, introducing noise in the action that a policy would take in such a state will significantly deteriorate performance (the return will be lower). Here we follow the same protocol of previous approaches~\cite{guo2021edge}, and we compare the policy's performance drop to the baseline methods when the 30 most critical states are attacked (i.e., whenever the agent reaches those states, its action is perturbed).

\begin{table}[]
\caption{\textbf{Win rate changes of the agent before/after attacks by following the protocol of EDGE~\cite{guo2021edge}.} We use the detected top 30 states as input to attack the policy. We report means and standard deviations over three random seeds. The reported results of all the baselines are from previous work~\cite{guo2021edge}. $\mathbf{s},\mathbf{a},\mathbf{y},\pi$ denote the state, action, return, and policy parameters, respectively.}. 
\centering
\footnotesize
\begin{tabular}{p{3.4cm}|p{1.3cm}|>{\centering\arraybackslash}p{2.3cm}}
\toprule
Method    & Input            & Win Rate Changes $\downarrow$           \\ \midrule
Rudder~\cite{arjona2019rudder}   & ($\mathbf{s}$, $\mathbf{a}$, $\mathbf{y}$)   & -19.93 $\pm$ 4.43          \\ 
Saliency~\cite{simonyan2013deep,smilkov2017smoothgrad,sundararajan2017axiomatic} & ($\mathbf{s}$, $\mathbf{a}$, $\mathbf{y}$)    & -30.33 $\pm$ 0.47          \\
Attention RNN~\cite{bahdanau2014neural} & ($\mathbf{s}$, $\mathbf{a}$, $\mathbf{y}$, $\pi$)   & -25.27 $\pm$ 1.79          \\ 
Rationale Net~\cite{lei2016rationalizing}   & ($\mathbf{s}$, $\mathbf{a}$, $\mathbf{y}$, $\pi$)   & -29.20 $\pm$ 4.24          \\ 
Edge  ~\cite{guo2021edge}   & ($\mathbf{s}$, $\mathbf{a}$, $\mathbf{y}$, $\pi$)   & -65.47 $\pm$ 2.90          \\ 
\rowcolor[HTML]{EFEFEF} 
Ours with single policy     & \textbf{($\mathbf{s}$, $\mathbf{y}$)} & \textbf{-77.67 $\pm$ 0.56} \\ 
\rowcolor[HTML]{C0C0C0}
Ours with multiple policies     & \textbf{($\mathbf{s}$, $\mathbf{y}$)} & \textbf{-85.90 $\pm$ 1.47} \\ \bottomrule
\end{tabular}
\vspace{-1.8em}
\label{tab:adv_attack}
\end{table}

Table \ref{tab:adv_attack} shows that our method outperforms the other techniques in the Atari-Pong environment, exhibiting the most significant changes in win rates, highlighting its efficacy in localizing critical states.
In particular, we achieve an 18.63\% improvement over the previous SOTA method Edge\cite{guo2021edge}, suggesting that the states identified by our Deep State Identifier are more crucial to achieve a high return. Note that the previous methods, such as Edge~\cite{guo2021edge}, are based on sequences of states and action pairs. Our method instead achieves higher performance by only observing a state sequence. In the real-world scenario, imaging systems can easily capture sequences of visual states, while actions are more difficult to collect, requiring special sensors or manual annotations. In other words, our method can work with pure visual information to achieve higher performance, resulting in flexibility toward various potential applications. Moreover, when different policies collect the training dataset, the proposed method can benefit from data diversity, inducing more satisfactory results (i.e., an 85.90 drop in winning performance). 

\begin{table}[]
\caption{\textbf{Win rate changes of the agent before/after attacks for different policies.} We assess whether our method, trained on trajectories generated by one or multiple policies, can accurately identify critical time points within a trajectory generated by another unseen policy. We consider three kinds of unseen policies, including different random seeds (seeds), different training steps (steps), and different network architectures (Arch.), to test the performance of our method against cross-policy challenges. We report mean and standard error over three random seeds. We attack the policy perturbing its action in the top 30 states detected. }
\label{tab:cross_attack}
\vspace{3pt}
\Large
\resizebox{.48\textwidth}{!}{
\begin{tabular}{l|c|c|c}
\toprule
                                         & Baseline  & Ours (Single) & Ours (Multi.)  \\ \midrule
In-Policy (baseline)                     &  54.88 $\pm$ 1.80                  & -77.67 $\pm$ 0.56  & \cellcolor[HTML]{EFEFEF} \textbf{-85.90 $\pm$ 1.47} \\ 
Cross-Policy (Seeds)        &  -63.32 $\pm$ 0.93                  & -30.67 $\pm$ 0.58  & \cellcolor[HTML]{EFEFEF} \textbf{-85.45 $\pm$ 0.86}\\ 
Cross-Policy (Steps)        &  -50.23 $\pm$ 1.21                  & -30.57 $\pm$ 1.01  & \cellcolor[HTML]{EFEFEF} \textbf{-83.72 $\pm$ 0.91} \\ 
Cross-Policy (Arch.) &      -49.85 $\pm$ 3.50                             &  -39.55 $\pm$ 2.38 &  \cellcolor[HTML]{EFEFEF} \textbf{-76.50 $\pm$ 3.11}      \\  \bottomrule
\end{tabular}
}
\vspace{-0.8em}
\end{table}

We then analyze the attack performance across different policies to test the robustness against policy shifts. In Table \ref{tab:cross_attack}, we set the baseline that attacks 30 states chosen randomly and attacks a policy that was never used to train our method. To ensure policy diversity for testing, we derive the policies with various random seeds, training steps, and network architectures. Compared with the baseline, our method cannot improve performance using a single policy, which indicates that a cross-policy protocol is challenging for adversarial attacks. However, when we increase the training data diversity by adding policies, we achieve a higher generalization, and the model's drop in performance improves from 49.85 to 76.50. A potential explanation is that each policy induces a specific distribution over the state space in the environment. Using different policies to collect data allows us to generalize to unseen policies and achieve more invariant representations of the policy behavior. Indeed, when the dataset can cover the distribution of states in the environment, our method generalizes to arbitrary unseen policies. We thereby achieve an environment-specific policy-agnostic solution for interoperability.

\begin{table}[]
\caption{\textbf{Performance of DQN with different adaptive step strategies on Atari-Seaquest.} We base the implementation on the Tianshou Platform~\cite{tianshou}. Our method effectively improves the performance of DQN. n-step stands for the lookahead steps.}
\label{tab:policy_improve}
\footnotesize
\resizebox{.475\textwidth}{!}{
\centering
\begin{tabular}{l|c}
\hline
Methods                                                          & Return $\uparrow$ $\pm$ St.d.     \\ \hline
PPO (time steps=5M)~\cite{schulman2017proximal}                 & 887.00 $\pm$ 4.36                    \\ 
SAC (time steps=5M)~\cite{haarnoja2018soft}                 & 1395.50  $\pm$ 339.34    \\ 
Rainbow (step=3,time steps=5M)  ~\cite{hessel2018rainbow}                   & 2168.50  $\pm$ 332.89   \\ \hline
DQN(time steps=10M)~\cite{mnih2013playing}                         & 3094.75 $\pm$ 1022.54   \\
DQN (n-step=random(1,5),time steps=5M)~\cite{sutton1988learning}                & 3250.25 $\pm$ 638.13    \\
\rowcolor[HTML]{EFEFEF} 
Baseline: DQN (n-step=5,time steps=5M) ~\cite{sutton1988learning}                         & 1987.00  $\pm$ 115.71   \\ 
DQN (n-step=12,time steps=5M)~\cite{sutton1988learning}                          & 1472.50  $\pm$  407.40  \\ 
DQN (n-step=grid search,time steps=5M)~\cite{sutton1988learning}                          & 3936.50  $\pm$  459.19  \\ \hline
SAC (time steps=25M)\cite{haarnoja2018soft}                         & 1444.00 $\pm$ 136.86   \\
Rainbow (time steps=25M)\cite{hessel2018rainbow}                         & 2151.25 $\pm$ 329.29   \\
DQN (time steps=25M)\cite{mnih2013playing}                         & 3525.00 $\pm$ 63.87   \\ \hline
HL based on Frequency (time steps=5M)\cite{mcgovern2001automatic,csimcsek2008skill} & 2477.00 $\pm$ 223.65                       \\ \hline
\rowcolor[HTML]{C0C0C0}
DQN + Ours (n-step$\leq$5,time steps=5M)  & \textbf{4147.25 $\pm$ 378.16}  \\ \hline
\end{tabular}
}
\vspace{-2em}
\end{table}

\subsection{Policy Improvement}
\label{sec:impr}
We show how our method can improve DRL policies. The experimental results in the previous sections demonstrate that our Deep State Identifier can efficiently identify critical states. Here we show how one can use these states to perform rapid credit assignment for policy improvement. In particular, we combine our method with the widely-used DQN~\cite{mnih2013playing} for multi-step credit assignment. The objective function of traditional Multi-step DQN\cite{hessel2018rainbow,sutton1988learning} is:

\begin{equation}
\begin{aligned}
\sum_{(s\timestep[j],a\timestep[j])\in\text{Rep.}} \Bigg[&  Q(s\timestep[j],a\timestep[j])  - \Bigg( \sum_{t=j}^{j+n-1} \gamma^{t-j} r\timestep[t] + 
\\
&\gamma^{n}\max_{a\timestep[{j+n}]}Q^{\rm T}(s\timestep[{j+n}],a\timestep[{j+n}]) \Bigg) \Bigg]^2,
\end{aligned}
\end{equation}
where $Q$ is the action-value function, i.e., a network predicting the expected return of the policy from a particular state-action pair, Rep. is the replay buffer, $Q^{\rm T}$ is a copy of $Q$, which is periodically synchronized with $Q$ to facilitate learning, $\gamma$ is the discount factor, and $a$ denotes an action.  

A recent study~\cite{wang2023highway} highlights the importance of varying the lookahead step $n$ in Multi-step DQN. Here we combine our method with Multi-step DQN by first identifying critical states and then dynamically setting lookahead steps to learn DQN. In other words, we set $n$ as the number of time steps from the state to the most critical state detected within a specific range. Here, we set the maximum lookahead step to 5. 

Table \ref{tab:policy_improve} presents preliminary results which illustrate that Multi-step DQN combined with our method improves the return of DQN from 1987.00 to 4147.25. Since our method effectively discovers states important for return prediction, our Deep State Identifier provides DQN with faster credit assignment, improving its performance. Moreover, our method performs slightly better than finely tuning the lookahead step $n$ using grid search.
Table \ref{tab:policy_improve} also includes improved versions of DQN~\cite{mcgovern2001automatic,csimcsek2008skill} for comparison. Our method outperforms all of them.

\section{Conclusion}

Our novel method identifies critical states from episodes encoded as videos. Its return predictor and critical state detector collaborate to achieve this.
When the critical state detector is trained, it outputs a soft mask over the sequence of states. This mask can be interpreted as the detector's belief in the importance of each state. Experimental results confirm that the generated belief distribution closely approximates the importance of each state. 
Our approach outperforms comparable methods for identifying critical states in the analyzed environments. It can also explain the behavioral differences between policies and improve policy performance through rapid credit assignment.
Future work will focus on applying this method to hierarchical RL and exploring its potential in more complex domains.

\section*{Acknowledgements}
We thank Dylan R. Ashley for his valuable comments and help to polish the paper.  This work was supported by the European Research Council (ERC, Advanced Grant Number 742870) and the SDAIA-KAUST Center of Excellence in Data Science and Artificial Intelligence (SDAIA-KAUST AI).  

{
\small
\bibliographystyle{ieee_fullname}
\bibliography{cite}
}

\clearpage
\appendix

This appendix provides the implementation details of our Deep State Identifier. In Section \ref{sec:impl}, we provide the pseudo-code for the Deep State Identifier, its network architecture, and the hyperparameters used during training. Then, Section \ref{sec:bench} discusses the datasets we collected and our experimental protocol. Finally, Section \ref{sec:exp} provides additional experimental results related to the ablation study and the comparison with EDGE \cite{guo2021edge} on MuJoCo.

\section{Implementation Details}
\label{sec:impl}
This section details our implementation of the proposed method. We implement our method and conduct our experiments using PyTorch~\cite{paszke2019pytorch}. All experiments were conducted on a cluster node equipped with 4 Nvidia Tesla A100 80GB GPUs.

The proposed method---while quite effective---is conceptually simple. The training pipeline can be written in 25 lines of pseudo-code:

\begin{lstlisting}[language=Python]
import torch as T
def cs_detector_train(input_states, labels):
    mask = cs_detector(input_states)
    loss_reg = lambda_r*T.linalg.norm(mask,ord=1)
    masked_states = mask * input_states
    output = return_predictor(masked_states)
    loss_sub = lambda_s*criterion(output,labels)
    reverse_mask = torch.ones_like(mask) - mask
    reverse_states = reverse_mask * input_states
    output_r = return_predictor(reverse_states)
    confused_label = torch.ones_like(output_r)*0.5 #binary classification case
    loss_vic = lambda_v * criterion(output_r,confused_label)
    loss_total = loss_reg + loss_sub + loss_vic
    loss_total.backward()
    optimizer_cs.step()
def return_predictor_train(input_states, labels):
    output = return_predictor(input_states)
    loss_d = criterion(output,labels)
    loss_d.backward()
    optimizer_return.step()
def main_train(input_states, labels):
    optimizer_cs.zero_grad()
    cs_detector_train(input_states, labels)
    optimizer_return.zero_grad()
    return_predictor_train(input_states, labels)
\end{lstlisting}

We use two potential network architectures in our work, 3DCNN~\cite{tran2015learning}, and CNN-LSTM~\cite{fukushima1980neocognitron,Hochreiter:1997:LSM:1246443.1246450,krizhevsky2017imagenet}, to implement our Deep State Identifier. Tables \ref{tab:3dcnn} and \ref{tab:cnn_lstm} show the specification of the corresponding architectures.  We use 3DCNN architecture in Table \ref{tab:sup_ablation_study} and employ LSTM structure in the other empirical studies.
\begin{table}[!h]
\centering
\caption{\textbf{The specification of the 3DCNN-based Neural Network adopted in this paper.} In-Norm refers to the Instance Normalization, 3D Conv. is the 3D convolutional Layer, and F.C. refers to the fully connected layer. In the last layer, the [return predictor/critical state detector] has a different architecture specified in the last column.}
\label{tab:3dcnn}
\resizebox{.475\textwidth}{!}{
\begin{tabular}{c|c|c|c|c|c}
\hline
3DCNN               & Channel                      & Filter & Stride  & In-Norm & Activation \\ \hline
3D Conv.       & 12 $\rightarrow$ 32          & (1,3,3)     & (1,2,2) & False         & Relu             \\ \hline
3D Conv.       & 32 $\rightarrow$ 64          & (1,3,3)     & (1,1,1) & True          & Relu             \\ \hline
3D Conv.       & 64 $\rightarrow$ 128         & (1,3,3)     & (1,2,2) & False         & Relu             \\ \hline
3D Conv.       & 128 $\rightarrow$ 128        & (1,3,3)     & (1,1,1) & True          & Relu             \\ \hline
3D Conv.       & 128 $\rightarrow$256       & (3,2,2)     & (1,1,1) & False         & Relu             \\ \hline
Avg Pooling & -                            & -           & -       & -             & -                \\ \hline
F.C.             & 256 $\rightarrow$ 512        & -           & -       & -             & -                \\ \hline
F.C.             & 512 $\rightarrow$ {[}2/12{]} & -           & -       & -             & {[}-/sigmoid{]}  \\ \hline
\end{tabular}
}
\end{table}
\begin{table}[!h]
\centering
\caption{\textbf{The specification of the CNN-LSTM Neural Network in this paper.} In the last layer, the critical state detector outputs a vector with the same length as the input (i.e., 256$\rightarrow$1). The return predictor estimates a scalar for the whole episode (i.e., 256 $\times$ length $\rightarrow$2)}
\label{tab:cnn_lstm}
\resizebox{.475\textwidth}{!}{
\begin{tabular}{c|ccc|cc}
\hline
CNN-LSTM    & \multicolumn{1}{c|}{Channel}                & \multicolumn{1}{c|}{Filter} & Stride     & \multicolumn{1}{c|}{In-Norm} & Activation \\ \hline
2D Conv.    & \multicolumn{1}{c|}{3 $\rightarrow$ 32}     & \multicolumn{1}{c|}{3}      & 2          & \multicolumn{1}{c|}{False}   & Relu       \\ \hline
2D Conv.    & \multicolumn{1}{c|}{32 $\rightarrow$ 64}    & \multicolumn{1}{c|}{3}      & 1          & \multicolumn{1}{c|}{True}    & Relu       \\ \hline
2D Conv.    & \multicolumn{1}{c|}{64 $\rightarrow$ 128}   & \multicolumn{1}{c|}{3}      & 2          & \multicolumn{1}{c|}{False}   & Relu       \\ \hline
2D Conv.    & \multicolumn{1}{c|}{128 $\rightarrow$ 128} & \multicolumn{1}{c|}{3}      & 1          & \multicolumn{1}{c|}{True}    & Relu       \\ \hline
2D Conv.    & \multicolumn{1}{c|}{128 $\rightarrow$ 256}  & \multicolumn{1}{c|}{2}      & 1          & \multicolumn{1}{c|}{False}   & Relu       \\ \hline
Avg Pooling & \multicolumn{1}{c|}{-}                      & \multicolumn{1}{c|}{-}      & -          & \multicolumn{1}{c|}{-}       & -          \\ \hline
            & \multicolumn{1}{c|}{Input}                  & \multicolumn{1}{c|}{Hidden} & Bi-Direct. & \multicolumn{2}{c}{Activation}            \\ \hline
LSTM        & \multicolumn{1}{c|}{256}                    & \multicolumn{1}{c|}{128}    & True       & \multicolumn{2}{c}{-}                     \\ \hline
F.C.        & \multicolumn{3}{c|}{{[}length$\times$256{]} $\rightarrow$ {[}2/length{]}}                      & \multicolumn{2}{c}{{[}-/sigmoid{]}}       \\ \hline
\end{tabular}
}
\end{table}

To train the critical state detector and return predictor, we use the Adam optimizer~\cite{kingma2014adam} with $\beta_1=0.9$ and $\beta_2=0.999$. The learning rate is set as $1\times10^{-4}$ and the weight decay is $1\times10^{-4}$. The input length of 3DCNN is 12 frames and is a partial observation ($7\times7$ pixels) of the environment~\cite{minigrid, chevalier2018babyai}. The remaining hyper-parameters $\lambda_s$, $\lambda_r$, and $\lambda_v$ are set to $1$, $5\times10^{-3}$ and $2$ respectively. 

\begin{figure}[h!]
    \centering
    \vspace{12pt}
    \includegraphics[width=0.48\textwidth]{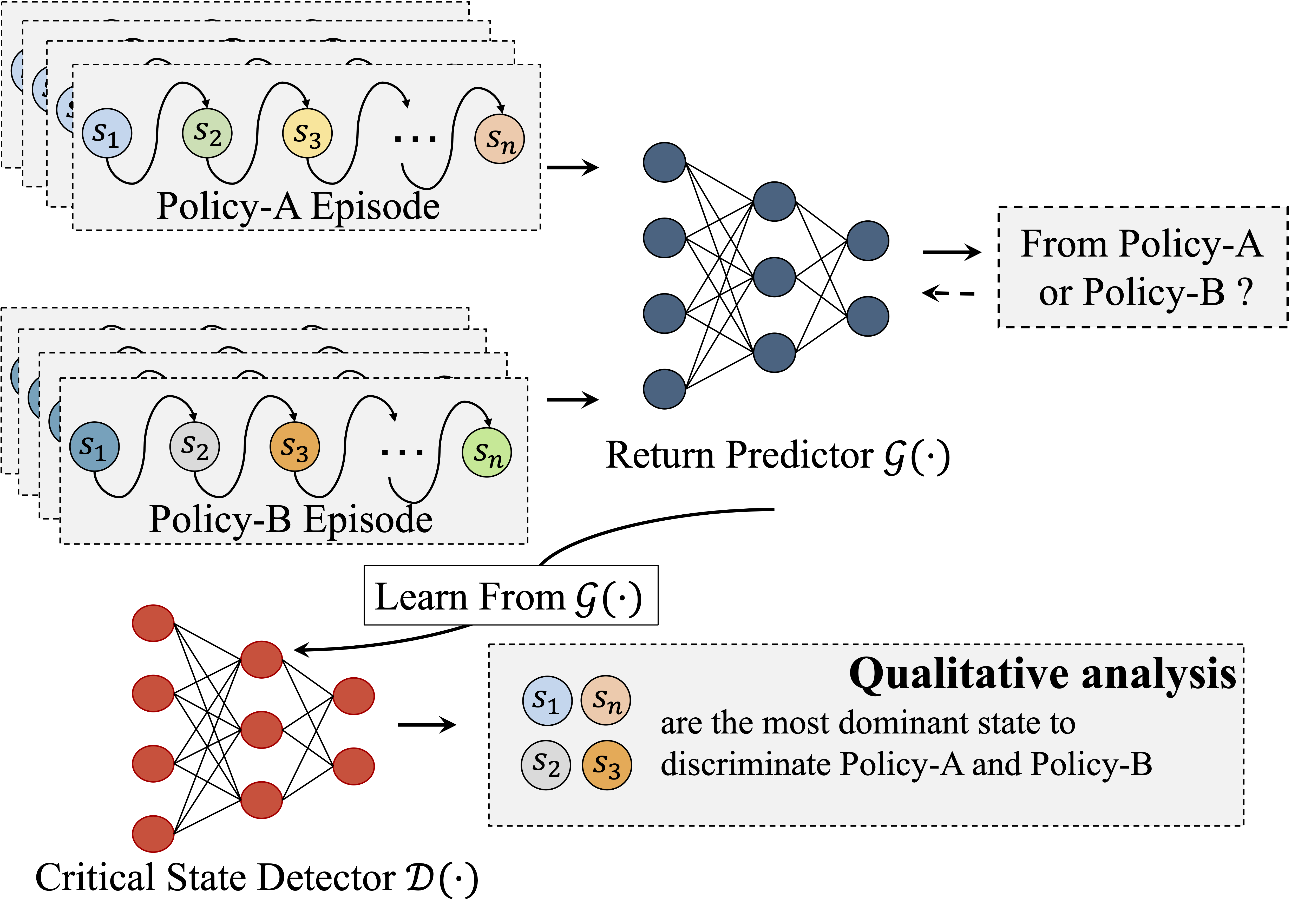}
   \caption{\textbf{Illustration of the Deep State Identifier for policy comparison.} We modify the return predictor as a binary classifier. Its training data comprises pairs $\{\mathbf{s_i,c_i}\}$, where $\mathbf{s_i}$ represents a trajectory and $c_i\in\mathbb{R}$ is a class label indicating whether it belongs to policy-\textit{A} or policy-\textit{B}.  By exploiting the return predictor, the critical state detector can directly localize the states that primarily explain the difference between policy-\textit{A} and policy-\textit{B}.}
   \label{fig:policy_comparison}
\end{figure}

Fig. \ref{fig:policy_comparison} shows how we can adapt the return predictor to find the critical frame that explains the difference in behavior between the two policies. We can train the return predictor to identify which of the two policies generates a specific trajectory.

\begin{figure*}[!h]
    \centering
    \includegraphics[width=0.98\textwidth]{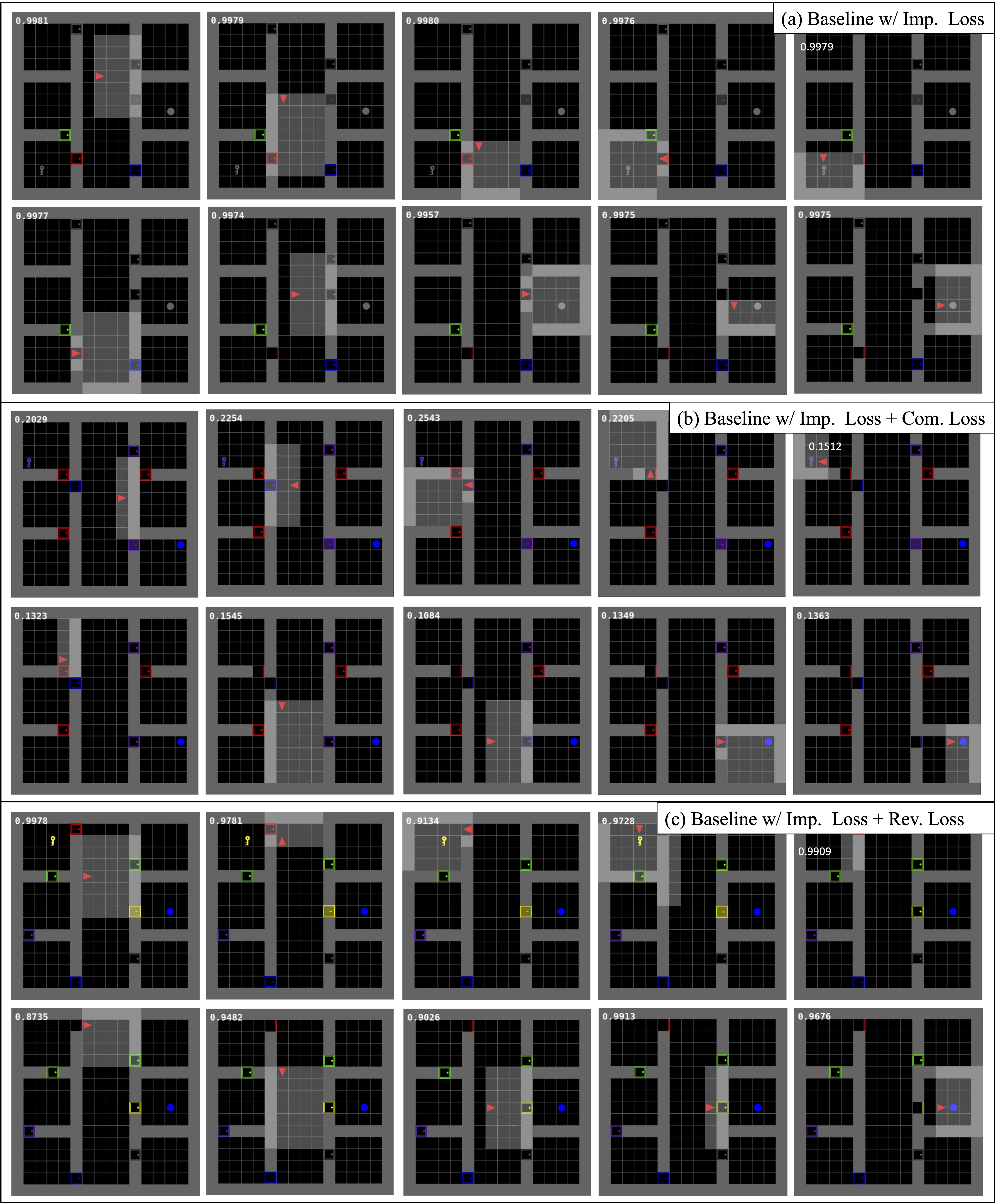}
   \caption{\textbf{Visualization of our method with different losses.} The number at the top-left corner indicates the confidence score predicted by the critical state detector, indicating whether the corresponding state is important. (a) Baseline trained with Importance Preservation loss; (b) Baseline with Importance Preservation loss and Compactness loss. (c) Baseline with Importance Preservation loss and Reverse loss. None of them can detect critical states effectively.  }
   \label{fig:badcase}
\end{figure*}

\begin{table*}[!h]
\centering
\caption{\textbf{Ablation study for the Deep State Identifier.} Clean Acc. refers to the accuracy of the return predictor in the test set; Masked Acc. is the accuracy of the return predictor with the input (critical states) detected by the critical state detector; R-Masked Acc. is the accuracy of the return predictor where the masked is inverted (non-critical states are treated as critical and vice versa); L1(Mask) and Var(Mask) are the L1 norm and the average variance of the output of the critical state detector respectively. }
\label{tab:sup_ablation_study}
\resizebox{1\textwidth}{!}{
\centering
\begin{tabular}{ccccc|c|c|c|c|c}
\toprule
\textit{Imp.  Loss}   &  \textit{Com. Loss} &\textit{Rev. Loss}     & 3DCNN & CNN-LSTM                                      & Clean Acc. (\%) $\uparrow$ & Masked Acc.(\%) $\uparrow$ & R-Masked Acc.(\%) $\downarrow$ & L1(Mask) $\downarrow$ & Var(Mask) $\uparrow$                            \\
\midrule
$\checkmark$   &$\times$ &$\times$ &$\checkmark$  & $\times$              & 90.07           & 90.07           & \textbf{44.57}             & 63.74    & 2 $\times$ $10^{-6}$ \\ 
$\checkmark$   &$\checkmark$ &$\times$ &$\checkmark$  & $\times$              & 91.45           & 87.71           & 89.28             & \textbf{8.79}     & 0.01                                \\ 
$\checkmark$   &$\times$ &$\checkmark$ &$\checkmark$  & $\times$          & 91.45           & 91.39           & 76.12             & 63.73    & 0.03                                \\ 
$\checkmark$   &$\checkmark$ &$\checkmark$ &$\checkmark$  & $\times$          & 90.78           & 89.45           & 64.55             & 57.35    & 0.04                                \\ 
\rowcolor[HTML]{EFEFEF} 
$\checkmark$   &$\checkmark$ &$\checkmark$ &$\times$  & $\checkmark$     & \textbf{98.66}           & \textbf{98.44}           & 55.58             & 41.05    & \textbf{0.12}                                  \\ \bottomrule 
\end{tabular}
}
\end{table*}

\begin{figure*}[!h]
    \centering
    \includegraphics[width=0.98\textwidth]{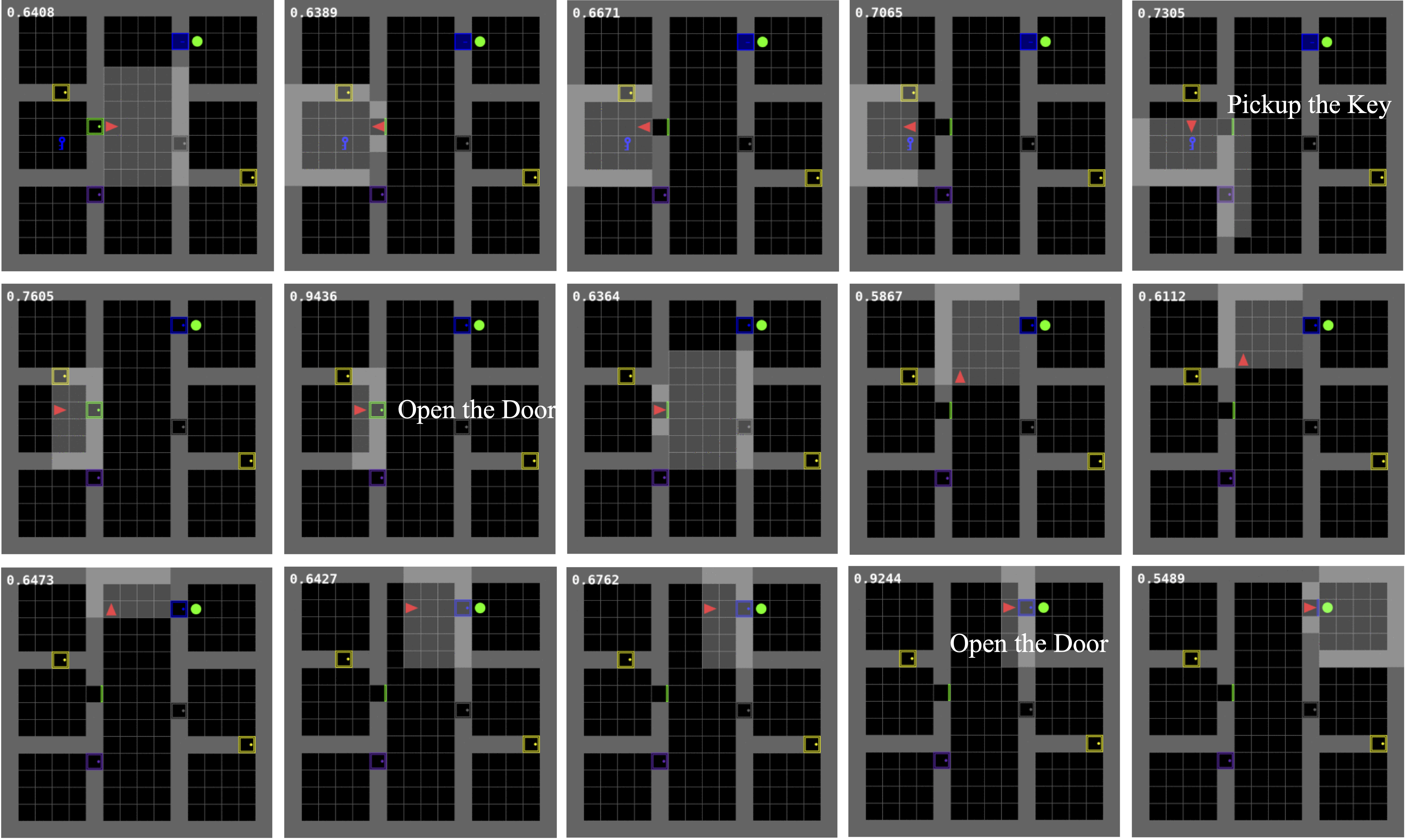}
   \caption{\textbf{Sampled observations from an episode collected by our method.} The number at the top-left corner indicates the confidence score predicted by the critical state detector, indicating whether the corresponding state is critical. Our method can localize the critical states effectively.}
   \label{fig:sup_ours_visualization}
\end{figure*}

\section{Experimental details}
\label{sec:bench}

\noindent
\textbf{Critical States Discovery.} We use a GridWorld environment (MiniGrid-KeyCorridorS6R3-v0) to collect a dataset (Grid-World-S) to test the accuracy of the critical state detector. Data is collected by acting in the environment using an optimal policy based on a depth-first search algorithm (DFS). Additional data is collected from a random-exploring policy. Since, in this environment, one can find critical states by visual inspection (they correspond to the states immediately before or after the action of opening doors or picking up keys), we can directly test the accuracy of the proposed method. We use the F1 score as a metric.

\noindent
\textbf{Policy Comparison by Critical States.} Here, we collect a dataset, \textit{Grid-World-M}, for our experiments on policy comparison. The labels in \textit{Grid-World-M} are the policies that collected the corresponding episode. We use two policies to collect data: Policy-A is the optimal policy used to collect \textit{Grid-World-S}, while Policy-B is an exploratory policy.

\noindent
\textbf{Efficient Attack using Critical States.}
Here we use adversarial attacks on Atari-Pong to validate whether the detected states are critical. Following the same protocol as Edge~\cite{guo2021edge}, we use a trained policy downloaded from \url{https://github.com/greydanus/baby-a3c} to collect the training data. We call the corresponding dataset \textit{Atari-Pong-S}. In particular, we collect 21500 episodes for training and 2000 for testing, and we fix the length of each episode as 200. We augment the input by randomly increasing or decreasing the length within 50 frames, and the padding value is set as 0. 
To validate the generalization of the proposed method for unseen policies, we then collect another dataset, denoted \textit{Atari-Pong-M}. We train policies with different seeds using the same implementation as Edge~\cite{guo2021edge} from \url{https://github.com/greydanus/baby-a3c}. In particular, 
 we use ten different policies to collect training data. In cross-policy (seeds), we use the trained policy on different random seeds to test the performance. In cross-policy (steps), we use the policy trained with 80M and 40M steps for training and testing our method, respectively. In cross-policy (Arch.), we change the architecture to make the setting more challenging. In particular, we train our method using a policy with 32 channels but test it by attacking a policy trained using 64 channels. The result in each case is collected by attacking the agent for 1500 episodes using three random seeds.

\noindent
\textbf{Policy Improvement.} We test the potential of our method to improve policy performance in the Atari-Seaquest environment. We first train the policies based on DQN following the implementation of Tianshou~\cite{tianshou}. Then we use the trained policies to collect a dataset called \textit{Atari-Seaquest-S}, consisting of 8000 trajectories for training and 2000 trajectories for testing. The average length of the trajectories is 2652.5, and the average return is 2968.6. We cut the trajectory into subsequences with 200 states for training. To stabilize the training, we equip an orthogonal regularization for our method. Considering the output of the predictor is a matrix, $\mathcal{M} \in \mathbb{R}^{b\times l}$ where $n$ refers to the batch size and $l$ is the length of $\mathbf{m}$, we drive the model to minimize the accumulation of $\mathcal{M} \mathcal{M}^T$. As the critical states of each trajectory are generally with different time steps, this regularization can benefit our approach. We train our Deep State Identifier on this video dataset and then test its effectiveness by re-training a new adaptive multi-step DQN from scratch, where the critical state detector adaptively determines the lookahead step. We use our trained critical state detector to determine the lookahead steps for rapid credit assignment during re-training. 

\begin{table}[t]
\caption{ \textbf{Sensitivity Analysis of the Deep State Identifier.} We show the F1 score $\uparrow$ of our method using different hyper-parameters on GridWorld-S datasets.}
\label{tab:hyper}
\setlength\tabcolsep{12pt}
\resizebox{.48\textwidth}{!}{
\begin{tabular}{c|c|c|
>{\columncolor[HTML]{EFEFEF}}c |c|c|
>{\columncolor[HTML]{EFEFEF}}c }
\toprule
$\lambda_r$($\times10^{-3}$) & 1  & 2.5 & 5  & 7.5 & 10  & \textbf{Variance} \\ \hline
F1 Score    & 76.69 & 78.44  & 80.28 & 78.26  & 76.44 & \textbf{1.39}     \\ \hline \midrule
$\lambda_s$ & 0.5   & 0.75   & 1     & 1.25   & 1.5   & \textbf{Variance} \\ \hline
F1 Score    & 76.68 & 77.50  & 80.28 & 78.18  & 78.83 & \textbf{1.22}     \\ \hline \midrule
$\lambda_v$ & 1.5   & 1.75   & 2     & 2.25   & 2.5   & \textbf{Variance} \\ \hline
F1 Score    & 77.77 & 77.04  & 80.28 & 78.76  & \textbf{83.78} & \textbf{2.39}     \\ \bottomrule
\end{tabular}
}
\end{table}

\begin{table}[h]
\caption{ \textbf{Win rate changes of the agent before/after attacks by following the protocol of EDGE \cite{guo2021edge}} We compare the methods on two MuJoCo environments: You-Should-Not-Pass game \cite{bansal2017emergent} (MuJoCo-Y) and Kick-And-Defend game \cite{bansal2017emergent} (MuJoCo-K). }
\label{tab:mujuco}
\setlength\tabcolsep{22pt}
\resizebox{.48\textwidth}{!}{
\begin{tabular}{l|c|c}
\toprule
 Method         & MuJoCo-Y        & MuJoCo-K        \\ \midrule
Rudder~\cite{arjona2019rudder}    & -32.53          & -21.80          \\ 
Saliency~\cite{simonyan2013deep,smilkov2017smoothgrad,sundararajan2017axiomatic}  & -29.33          & -37.87          \\ 
Attention RNN~\cite{bahdanau2014neural} & -33.93          & -41.20          \\ 
Rationale Net~\cite{lei2016rationalizing}    & -30.00          & -7.13           \\ 
Edge  ~\cite{guo2021edge}      & -35.13          & -43.47          \\ \midrule
\rowcolor[HTML]{EFEFEF} 
Ours      & \textbf{-45.10} & \textbf{-48.03} \\ \bottomrule
\end{tabular}
}
\end{table}

\section{Experimental Results}
\label{sec:exp}
To justify the effectiveness of the proposed method, we carry out some additional visualization and analysis. Table \ref{tab:sup_ablation_study} shows some statistics of the output of the critical state detector and return predictor. We observe that the identified states are few (the L1 Norm is low), and the output of the return predictor does not change when it ignores non-critical states. If instead, the return predictor observes only states identified as non-critical, then the performance is much lower. These results further validate the effectiveness of the proposed method.
We provide additional visualization of the performance of our method when using different losses for the critical state detector. The results are consistent with our empirical studies. In particular, Fig.~\ref{fig:badcase}(a) shows that when using only the importance preservation loss, all the states are considered critical. When adding only the compactness loss (see Fig.~\ref{fig:badcase}(b)) or the reverse loss (see Fig.~\ref{fig:badcase}(c)), the performance is still not satisfactory. The proposed method can precisely detect the critical states only when using all three losses. Indeed, as shown in Fig.~\ref{fig:sup_ours_visualization}, our method correctly outputs high confidence when the agent observes critical states (0.73, 0.94, and 0.92) and low confidence (0.6) otherwise. 

\subsection{Non-Vision Environment}
We also tested the performance in non-vision environments \cite{bansal2017emergent} and compared our method with the same methods in Table \ref{tab:adv_attack}. As shown in Table \ref{tab:mujuco}, our method achieves a win rate change of \textbf{-45.10} on the MuJoCo~\cite{todorov2012mujoco} environment You-Shall-Not-Pass game, surpassing the performance of EDGE (-35.13) by 28.38\%. In the Kick-and-Defense environment, our method achieves a win rate change of \textbf{-48.03}, outperforming EDGE (-43.47) by 10.49\%. The consistent improvement indicates that our method exhibits strong robustness in non-vision environments. 

\subsection{Sensitivity analysis}
We evaluate the performance of our method using different values of hyperparameters $\lambda_r$,$\lambda_s$, and $\lambda_v$. Table \ref{tab:hyper} shows that our algorithm has moderate sensitivity to hyperparameters when their values are within a specific range. For example, given $\lambda_s \in [0.5,1.5]$, the performance variance is only 1.22, indicating stable performance. To determine the optimal hyperparameters, we searched a range of values. The best hyperparameters found in GridWorld-S were then used in all other environments. 

\end{document}